\documentclass[letterpaper]{article} 
\usepackage{aaai23}  
\usepackage{times}  
\usepackage{helvet}  
\usepackage{courier}  
\usepackage[hyphens]{url}  
\usepackage{graphicx} 
\usepackage{natbib}  
\usepackage{caption} 
\usepackage{algorithm}
\usepackage{algorithmic}

\usepackage{times}
\usepackage{epsfig}
\usepackage{graphicx}
\usepackage{amssymb}
\usepackage{CJKutf8}
\usepackage{adjustbox}
\usepackage{subfigure}

\usepackage{multirow}
\usepackage{hhline}
\usepackage{makecell}
\usepackage{xcolor}
\usepackage{color}
\usepackage{amsmath}
\usepackage{newfloat}
\usepackage{listings}
\DeclareCaptionStyle{ruled}{labelfont=normalfont,labelsep=colon,strut=off} 
\floatstyle{ruled}
\newfloat{listing}{tb}{lst}{}
\floatname{listing}{Listing}
\title{MGTANet: Encoding Sequential LiDAR Points Using Long Short-Term Motion-Guided Temporal Attention for 3D Object Detection}

\author{
    Junho Koh\equalcontrib,
    Junhyung Lee\equalcontrib,
    Youngwoo Lee,
    Jaekyum Kim,
    Jun Won Choi\thanks{Corresponding author.}
}
\affiliations{
    Hanyang University\\
    \{jhkoh, junhyunglee, youngwoolee, jkkim\}@spa.hanyang.ac.kr, junwchoi@hanyang.ac.kr
}

\usepackage{bibentry}

\begin{document}

\maketitle

\begin{abstract}
Most scanning LiDAR sensors generate a sequence of point clouds in real-time. While conventional 3D object detectors use a set of unordered LiDAR points acquired over a fixed time interval, recent studies have revealed that substantial performance improvement can be achieved by exploiting the {\it spatio-temporal context} present in a sequence of LiDAR point sets. In this paper, we propose a novel 3D object detection architecture, which can encode LiDAR point cloud sequences acquired by multiple successive scans. The encoding process of the point cloud sequence is performed on two different time scales. We first design a {\it short-term motion-aware voxel encoding} that captures the short-term temporal changes of point clouds driven by the motion of objects in each voxel. We also propose {\it long-term motion-guided bird's eye view (BEV) feature enhancement} that adaptively aligns and aggregates the BEV feature maps obtained by the short-term voxel encoding by utilizing the dynamic motion context inferred from the sequence of the feature maps. The experiments conducted on the public nuScenes benchmark demonstrate that the proposed 3D object detector offers significant improvements in performance compared to the baseline methods and that it sets a state-of-the-art performance for certain 3D object detection categories. Code is available at 
\textcolor{magenta}{https://github.com/HYjhkoh/MGTANet.git}.

\end{abstract}

\section{Introduction}

3D object detectors detect, localize, and classify objects in a 3D coordinate system. Notably, 3D object detection is essential in various robotic and autonomous driving applications. Accordingly, to date, various LiDAR-based 3D object detectors have been proposed \cite{second, pointpillars, voxelnet, voxelrcnn, pointrcnn, 3dssd, pointgnn, fastpointrcnn, STD, pvrcnn, sa-ssd}. In these works, 3D object detection was performed based on a single set of LiDAR point clouds acquired from a fixed number of laser scans. The point data in each set are unordered, and only the geometrical distributions of points are used to extract the features. In several robotics applications, LiDAR sensors stream point cloud sequences in real time through continuous scanning. However, existing 3D object detection methods do not utilize the temporal distribution of sequential LiDAR points, leaving room for improving the detection performance. 

\begin{figure}[t]
    \centering
    \begin{subfigure}[]
    {
        \label{sm_vfe_motivation}
        \includegraphics[width=0.97\columnwidth]{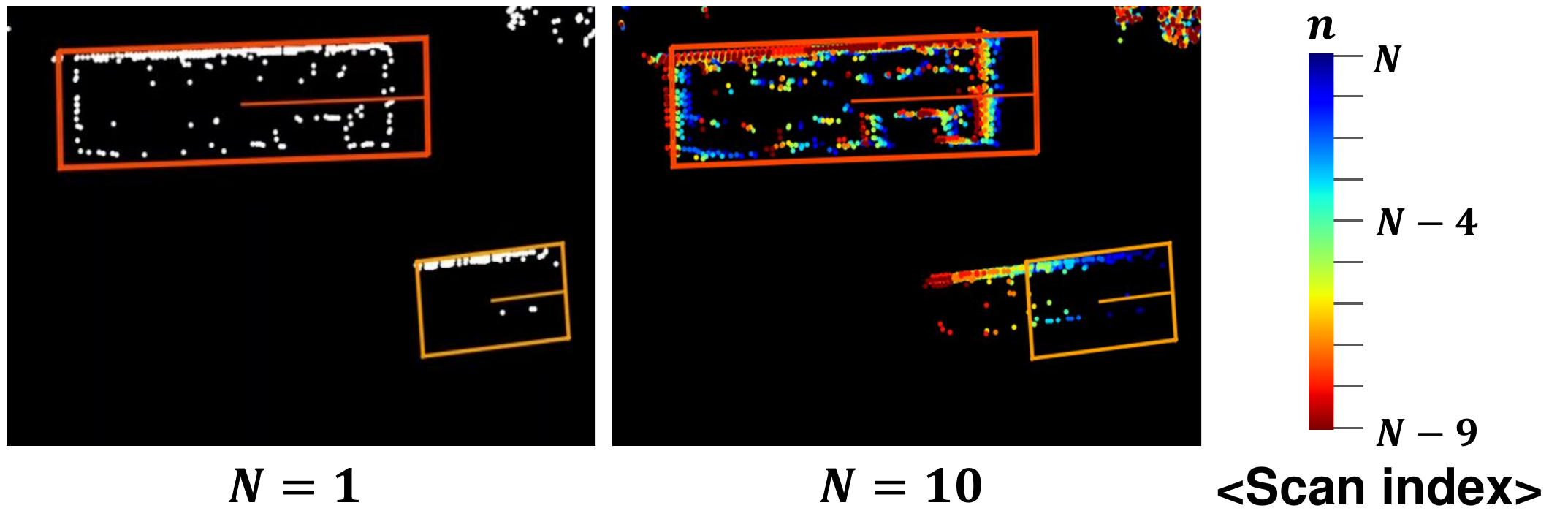}
    }
    \end{subfigure}
    \vspace{5mm}
    \begin{subfigure}[]
    {
        \label{mg_fam_motivation}
        \includegraphics[width=0.97\columnwidth]{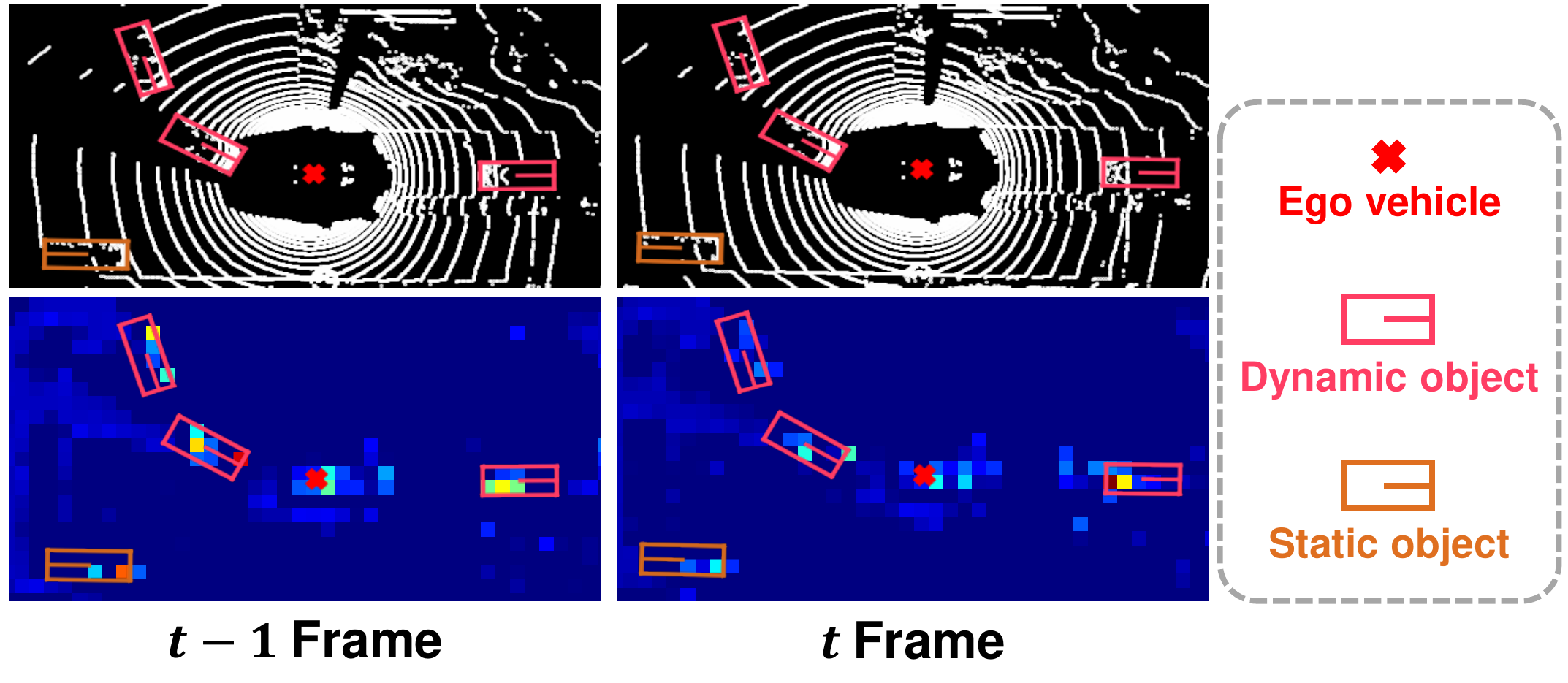}
    }
    \end{subfigure}
    \hspace{5mm}
    \caption { Visualization of (a) LiDAR points acquired over multiple scans (1 scan versus 10 scans) and (b) bird's-eye-view (BEV) feature maps obtained from the adjacent frames. 
    }
    \label{motivation}
\end{figure}

In the literature, there exist numerous relevant studies. Similar problems have been considered in {\it video object detection}, which detects objects using multiple adjacent video frames  \cite{D&T,fgfa,manet,rdn,selsa,psla,vod-mt,mega,tm-vod}. The primary interest of these studies focused on enhancing visual features using a sequence of video frames. Recently, this study has been extended to LiDAR-based 3D object detection \cite{3dvodlstm,3dvid,tctr,velocitynet,3dman}. In this study, the 3D object detectors that consume the temporal sequence of point clouds are called {\it 3D object detection with point cloud sequence} (3D-PCS). 
Several 3D-PCS methods have employed various sequence encoding models to aggregate the intermediate features obtained from each set of points \cite{3dvodlstm,3dvid,tctr,velocitynet,3dman}. The method in \cite{3dvodlstm} was based on a ConvLSTM model \cite{convlstm}  modified to capture a temporal structure in a sequence of point features. 3DVID \cite{3dvid} incorporated an attentive spatio-temporal memory in a ConvGRU model \cite{ConvGRU}  to enable spatio-temporal reasoning across a long-term point cloud sequence. VelocityNet \cite{velocitynet} used deformable convolution to generate  spatio-temporal features based on the velocity patterns of dynamic objects. 
3D-MAN \cite{3dman} attempted proposal-level feature aggregation using a transformer architecture.

In this paper, we propose a new 3D object detector, referred to as {\it Motion-Guided Temporal Attention Network} (MGTANet), designed to encode a finite-length sequence of LiDAR point clouds. The proposed MGTANet finds a representation of sequential point sets over two different time scales.

First, $N$ sets of points acquired by successive LiDAR scans constitute a single {\it frame} and are encoded by {\it short-term motion-guided BEV feature extraction}. Figure \ref{motivation} (a) illustrates the example of the point sets acquired by multiple scans. The points from different scanning grids can be observed to exhibit certain temporal patterns due to the motion of objects. These patterns can be used as contextual cues to enhance the features of objects. While most existing 3D object detectors merge these sets of points without considering their temporal context, we devise {\it short-term motion-aware voxel feature encoding} (SM-VFE) to arrange the sets of points in the scanning order and perform sequence modeling in the latent motion embedding space. The voxel features produced by the short-term voxel encoding are then transformed into {\it bird's-eye-view} (BEV) domain features using a standard {\it convolutional neural network} (CNN) backbone.

Second, $K$ BEV feature maps obtained by the short-term voxel encoding over $K$ successive frames are aggregated through {\it long-term motion-guided BEV feature enhancement}.  3D motion of objects causes dynamic changes in BEV features over frames (see Figure \ref{motivation} (b)), so these features should be aligned to boost the effect of feature aggregation. Motivated by the idea that the motion context provides a valuable clue for temporal feature alignment, we propose the {\it motion-guided deformable alignment} (MGDA) that extracts the motion features from two adjacent multi-scale BEV features and uses them to determine the position offsets and weights of the deformable masks. We also propose a novel {\it spatio-temporal feature aggregation} (STFA) that combines the aligned BEV features via spatio-temporal deformable attention. While the existing deformable DETR \cite{def-detr} cannot adaptively apply a deformable mask to each BEV feature map due to structural limitations, we introduce the concept of {\it derivative queries} to utilize the relationship with the feature map of interest in aggregating the adjacent BEV feature maps. This novel structure allows multiple BEV feature maps to be weighted and aggregated in an effective and computationally-efficient manner.

Our experiment results on a widely used public nuScenes dataset \cite{nuscenes} confirm that the proposed MGTANet outperforms existing 3D-PCS methods by significant margins and set a  state-of-the-art performance in some evaluation metrics in the benchmark.

The key contributions of our study are summarized as follows.
\begin{itemize}
    \item We propose a new 3D object detection architecture, MGTANet that exploits the spatio-temporal information in point cloud sequences both in short-term and long-term time scales.
    
    \item We design an enhanced voxel encoding architecture that performs sequence modeling by considering LiDAR scanning orders in a short sequence. In order to model points acquired by successive LiDAR scans in each voxel, a latent motion feature is augmented to each voxel representation. To the best of our knowledge, voxel encoding that accounts for temporal point distribution has not been introduced in the previous literature.
    
    \item We present a long-term feature aggregation method to find the representation of multiple BEV feature maps that dynamically vary over longer time scales. Our evaluation shows that motion context information extracted from the adjacent BEV feature maps plays a pivotal role in finding better representation of the sequential feature maps. Combination of both short-term and long-term encoding of LiDAR point cloud sequences offers performance gains up to 5.1 \% in mean average prediction (mAP) and 3.9 \% in nuScenes detection score (NDS) over CenterPoint baseline \cite{centerpoint} on the nuScenes 3D object detection benchmark. 

\end{itemize}

\begin{figure*}[t]
	\centering
        \centerline{\includegraphics[width=1.0\textwidth]{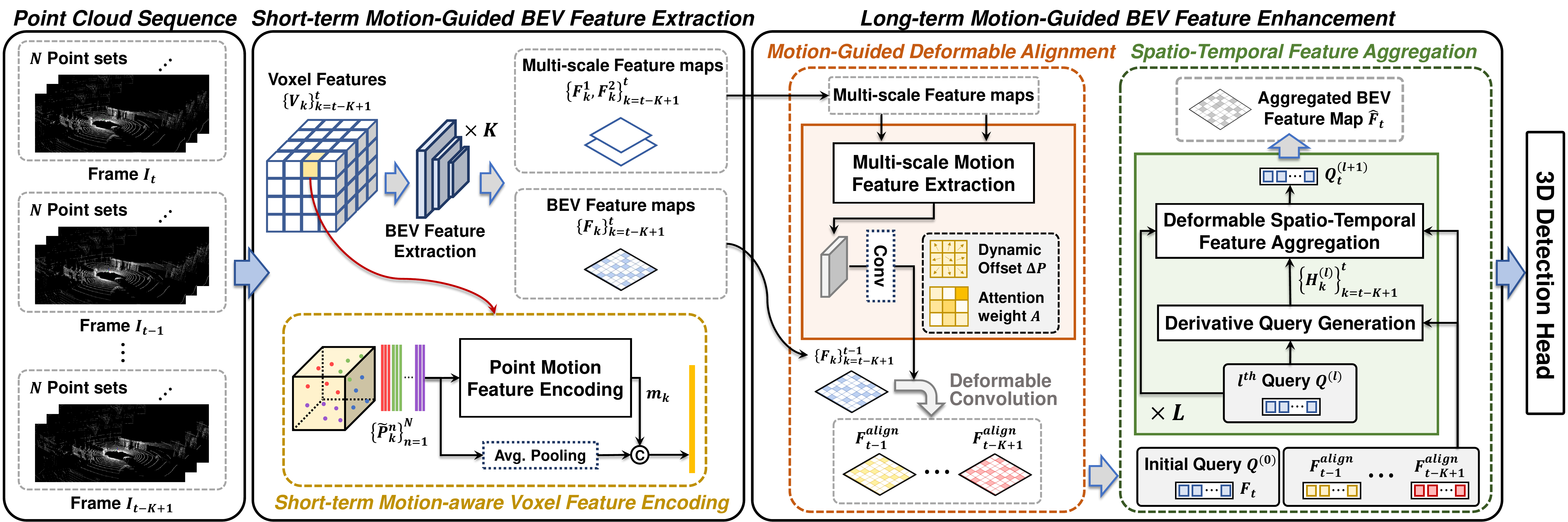}}
    	\caption {\textbf{Overall architecture of the proposed MGTANet.} MGTANet comprises three main blocks. SM-VFE performs voxel encoding based on a LiDAR point cloud sequence acquired over multiple LiDAR scans in a short sequence. MGDA then aligns the previous BEV feature maps to the current BEV feature map. Finally, STFA aggregates only the relevant parts of the aligned feature maps using a deformable cross-attention mechanism.}
	\label{overall}
\end{figure*}

\section{Related Work}
\subsection{3D Object Detection with a Single Point Set}
LiDAR-based 3D object detectors can be roughly categorized into  grid encoding-based methods \cite{second, voxelnet, pointpillars, voxelrcnn}, point encoding-based methods \cite{pointrcnn, 3dssd, pointgnn},  and hybrid methods \cite{fastpointrcnn, STD, pvrcnn, sa-ssd}. Among these, grid encoding-based methods organize the irregular point clouds using  regular grid structures such as voxels \cite{voxelnet} or pillars \cite{pointpillars} and  extract  the volumetric representations.  
In contrast, point encoding-based methods retain geometrical information of point clouds and extract point-wise features using PointNet \cite{pointnet} or PointNet++ \cite{pointnet++}. 
Moreover, some approaches take advantage of the merits of both these methods. 
In our study, the grid encoding-based methods are adopted as a baseline detector since they are more prevalent in training large-scale datasets such as nuScenes with state-of-the-art detection performance.

\subsection{3D Object Detection with Point Cloud Sequence}
The performance of 3D object detection methods can be improved by exploiting the temporal information in the LiDAR point cloud sequences. To date, several 3D-PCS methods have been proposed \cite{3dvodlstm, 3dvid, tctr, velocitynet, 3dman}. These methods explored various ways to find the representation of time-varying features obtained from multiple point sets. The method proposed in \cite{3dvodlstm} encoded temporal LiDAR features using long short term memory (LSTM) and 3DVID \cite{3dvid} realized an improved sequence modeling ability using an attentive spatio-temporal Transformer GRU. 
TCTR \cite{tctr} modeled temporal-channel information of multiple frames and decoded spatial-wise information using a transformer architecture. VelocityNet \cite{velocitynet} aligned temporal feature maps using deformable convolution driven by a motion map obtained from the velocities of objects. 3D-MAN \cite{3dman} stored the proposals and features obtained from a fast single-frame detector in a memory bank and fused them using a multi-view alignment and aggregation module.

The proposed MGTANet differs from the aforementioned methods in that it presents a holistic approach to process the temporal information of the point cloud sequences. Within point sets that exhibit only slight motion on a short-term scale, the proposed method captures the temporal distribution of points through voxel encoding. 
The proposed approach also considers the novel motion-guided feature alignment approach to actively adapt to dynamic feature changes occuring over longer time scales.

\section{Proposed Method}
In this section, we present the details of the proposed 3D object detector based on point cloud sequences.

\subsection{Overview}
The overall architecture of the proposed MGTANet is depicted in Figure \ref{overall}. The proposed 3D-PCS method consists of three main blocks; 1) {\it short-term motion-aware voxel feature encoding} (SM-VFE) 2) {\it motion-guided deformable alignment} (MGDA), and 3) {\it spatio-temporal feature aggregation} (STFA). Herein, SM-VFE is performed in a short sequence to encode enhanced voxel representations, whereas both MGDA and STFA are performed in long sequences to extract spatio-temporal BEV features under the guidance of motion information.

We assume that a LiDAR sensor generates a sequence of point clouds as it scans. The point clouds $P_k^n$ denote the point set acquired from the $n$th scanning step and the $k$th frame.  The $k$th frame consists of $N$ consecutive point sets, i.e., $I_k=\{P_k^n \}_{n=1}^{N}$. The ego-motion compensation is applied to the point sets within each frame \cite{nuscenes}. MGTANet uses a total of $N \times K$ point sets in the latest $K$ frames $\{I_k\}_{k=t-K+1}^{t}$ as inputs and produces the object detection results for the $t$th frame, where $t$ denotes the index for the frame of interest.

The $N$ point sets in each frame are encoded using the SM-VFE. First, these $N$ point sets are voxelized by a pre-defined grid size. Each voxel contains the points that belong to different point sets $P_k^1,P_k^2, ..., P_k^N$  and exhibits a particular motion pattern over $N$ point sets. The proposed SM-VFE extracts a voxel embedding vector that models the temporal distribution of such points obtained from each voxel. 
The voxel features $\{\textbf{\textit{V}}_{k}\}_{k=t-K+1}^{t}$ generated by SM-VFE are further encoded by a sparse 3D CNN backbone to produce the $K$ BEV feature maps $\{\textbf{\textit{F}}_{k}\}_{k=t-K+1}^{t}$ \cite{second}.

\begin{figure*}[t]
    
    \centering
    \begin{subfigure}[\textbf{Multi-scale motion feature extraction}]{\includegraphics[height=75mm]{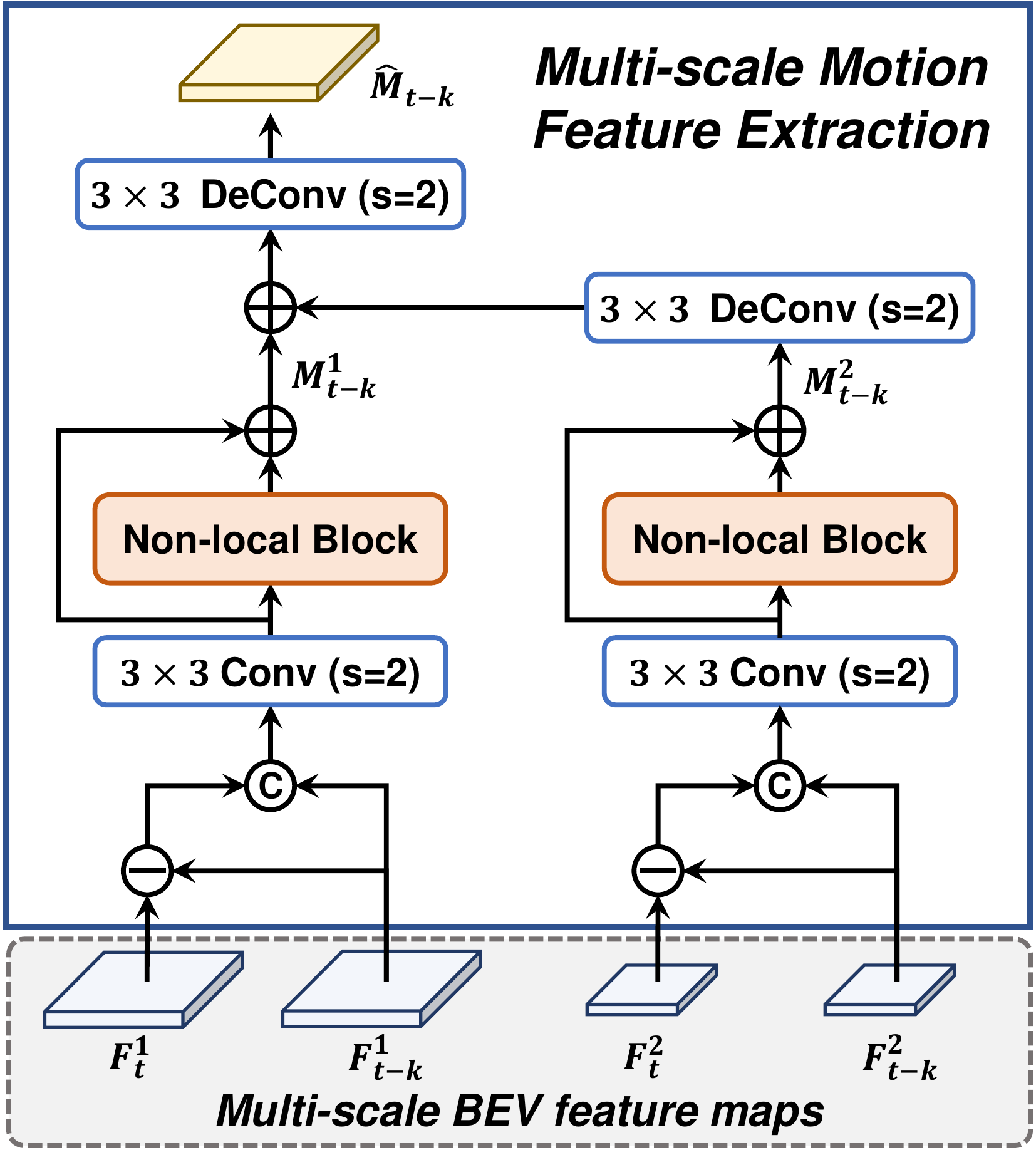}}
    \end{subfigure}
    \hspace{5mm}
    \begin{subfigure}[\textbf{Deformable spatio-temporal feature aggregation}]{\includegraphics[height=75mm]{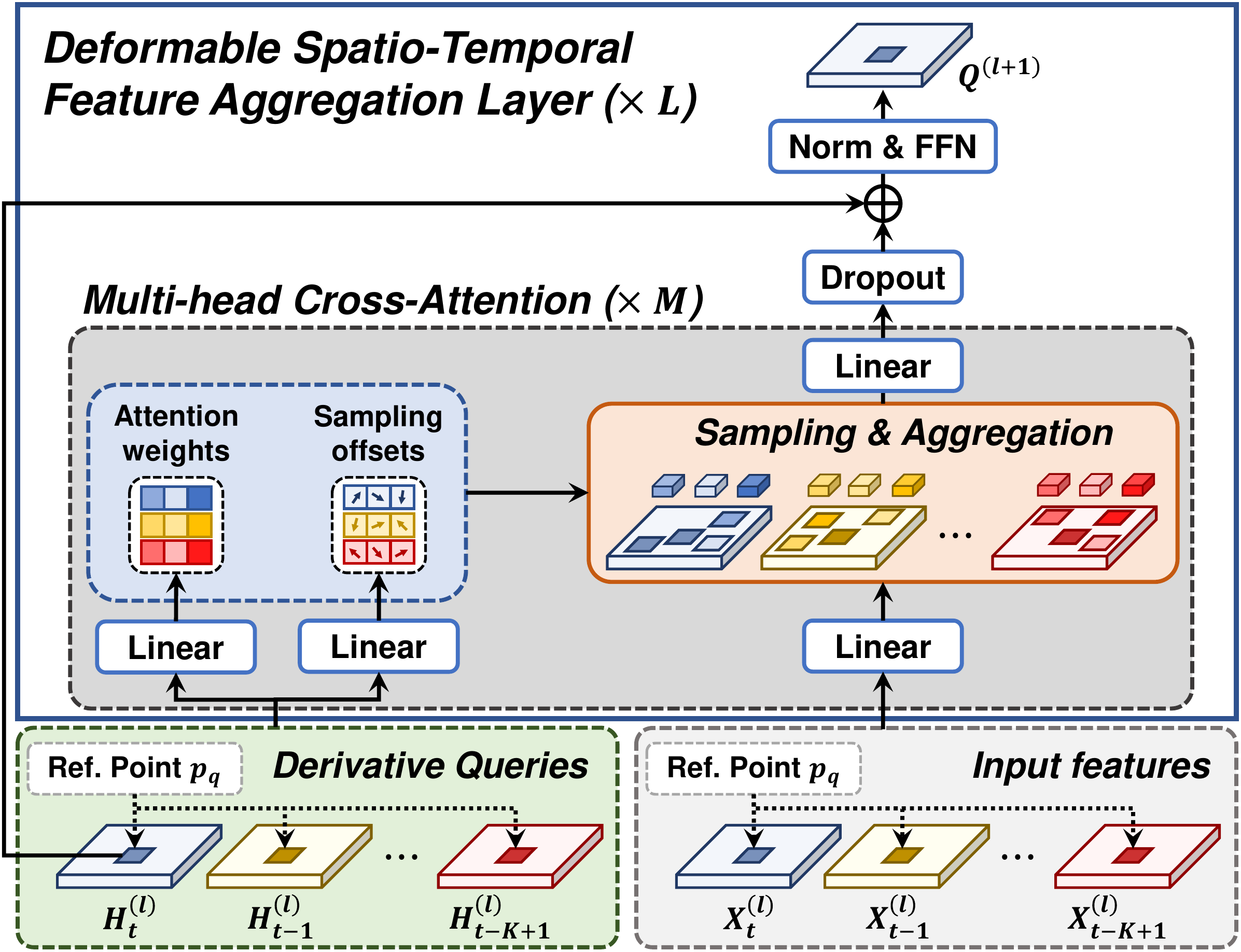}}
    \end{subfigure}
    \caption {\textbf{Structure of sub-modules in  long-term motion-guided BEV feature enhancement model} (a) {\it Multi-scale motion feature extraction} module extracts the motion context feature with multi-scale BEV features. (b) {\it Deformable spatio-temporal feature aggregation} module implements multi-head cross-attention to aggregate the adjacent BEV feature maps. 
    }
    \label{long_term_model}
\end{figure*}

Next, the BEV feature maps $\{\textbf{\textit{F}}_{k}\}_{k=t-K+1}^{t}$ are aggregated over a longer time scale. Before these features are aggregated, they are spatially aligned to improve the effect of feature aggregation.  First, MGDA aligns the previous $K-1$ BEV feature maps $\{\textbf{\textit{F}}_{k}\}_{k=t-K+1}^{t-1}$ to the current BEV feature map $\textbf{\textit{F}}_{t}$ under the guidance of contextual motion information. MGDA performs deformable convolution \cite{def-conv} to align each BEV feature map. The motion features extracted from multi-scale BEV feature maps are used to determine the sampling offsets and attention weights of the deformable masks. MGDA produces the aligned BEV features $\textbf{\textit{F}}_{t-1}^{align},...,\textbf{\textit{F}}_{t-K+1}^{align}$.
Then, STFA aggregates the $K$ BEV feature maps $\textbf{\textit{F}}_{t},\textbf{\textit{F}}_{t-1}^{align},...,\textbf{\textit{F}}_{t-K+1}^{align}$ using the deformable cross-attention model modified to enable spatio-temporal attention.
Finally, STFA produces the aggregated feature map, $\hat{\textbf{\textit{F}}}_t$, which is further processed by  the 3D detection head to output the final 3D detection results.

\subsection{Short-term Motion-aware Voxel Feature Encoding}
SM-VFE encodes LiDAR points considering the geometrical change occurring over time in each voxel. First, the $N$ point sets $\{P_k^n \}_{n=1}^{N}$ in the $k$th frame are merged and voxelized using a voxel structure \cite{voxelnet} or a pillar structure \cite{pointpillars}. Then, each voxel contains the $N$ point sets $\{\tilde{P}_k^n \}_{n=1}^{N}$ sub-sampled from $\{P_k^n \}_{n=1}^{N}$.  Let the cardinality of  $\tilde{P}_k^n$ be $N_k^n$ and the $i$th element of $\tilde{P}_k^n$ be  $\tilde{p}_k^n(i)$. The point entity $\tilde{p}_k^n(i)$ contains the $(x,y,z)$ 3D coordinates, reflectance, and time lag of a LiDAR point. We perform averaging operation over the elements of each point set $\tilde{P}_k^n$ as
\begin{align}
    \bar{p}_{k}^{n} = \frac{1}{N_{k}^n}\sum_{i=1}^{N_{k}^n} \tilde{p}_{k}^{n}(i).
\end{align}
If $\tilde{P}_{k}^{n}$ is empty, $\bar{p}_{k}^{n}$ is set to a zero vector of the same length. 
Next, we encode the sequence of $N$ points $\{\bar{p}_{k}^{n}\}_{n=1}^{N}$. For this, {\it differential encoding} is applied to represent the motion with respect to the last point $\bar{p}_{k}^{N}$, i.e., 
$D_k = \{\bar{p}_{k}^{N}-\bar{p}_{k}^{n}\}_{n=1}^{N-1}$. Each motion vector $(\bar{p}_{k}^{N}-\bar{p}_{k}^{n})$ passes through the fully-connected ($\mathrm{FC}$) layer, which is followed by the channel-wise attention ($\mathrm{CWA}$) \cite{senet}, as given below
\begin{align}
    q_{k}^{n} = \mathrm{CWA}(\mathrm{FC}(\bar{p}_{k}^{N}-\bar{p}_{k}^{n})).
\end{align}
Finally, the sequence $\{q_{k}^{n}\}_{n=1}^{N-1}$ is concatenated and encoded by an additional $\mathrm{FC}$ layer 
\begin{align}
    m_{k} = \mathrm{FC}(\mathrm{Concat}(\{q_{k}^{n}\}_{n=1}^{N-1})).
\end{align}
This motion feature $m_{k}$ is concatenated to the original voxel features computed by VoxelNet \cite{voxelnet} or PointPillars \cite{pointpillars}. These motion-aware voxel features are delivered to a conventional sparse 3D CNN backbone for further encoding \cite{second}.

\subsection{Motion-Guided Deformable Alignment}
MGDA aligns the adjacent BEV feature maps $\{\textbf{\textit{F}}_{k}\}_{k=t-K+1}^{t}$ using motion-guided deformable convolution.
For feature alignment, MGDA applies a deformable convolution \cite{deformV2} to each of the previous feature maps $\{\textbf{\textit{F}}_{k}\}_{k=t-K+1}^{t-1}$. The deformable mask applied to $\textbf{\textit{F}}_{t-k}$ is determined by the motion features computed based on the following two features $\textbf{\textit{F}}_{t-k}$ and $\textbf{\textit{F}}_{t}$. 

Figure \ref{long_term_model} (a) presents the structure of the motion feature extraction block.
To extract the motion features, we consider the multi-scale features $\textbf{\textit{F}}_{t-k}^{1}, \cdots, \textbf{\textit{F}}_{t-k}^{S}$ obtained from the intermediate  layers of the network branch for producing $\textbf{\textit{F}}_{t-k}$. Similarly, 
the multi-scale features $\textbf{\textit{F}}_{t}^{1}, \cdots, \textbf{\textit{F}}_{t}^{S}$ can be obtained from the intermediate layers for $\textbf{\textit{F}}_{t}$. These multi-scale features aid in representing motion at different scales and granularities. We only consider two scales $S=2$ in our implementation. For each scale, the motion features are obtained from 
\begin{align}
    \tilde{\textbf{\textit{M}}}^s_{t-k} &= \mathrm{Conv}_{3\times3}([\textbf{\textit{F}}^{s}_{t-k}, (\textbf{\textit{F}}^{s}_{t}-\textbf{\textit{F}}^{s}_{t-k})]) \\    
    \textbf{\textit{M}}^s_{t-k} &= \tilde{\textbf{\textit{M}}}^s_{t-k} + \mathrm{NLblock}(\tilde{\textbf{\textit{M}}}^s_{t-k}), 
\end{align}
 where $s \in \{1,2\}$ denotes the scale index, $[\cdot,\cdot]$ denotes the channel-wise concatenation, and  $\mathrm{NLblock}$ denotes the non-local block \cite{nonlocal}. The features $\textbf{\textit{M}}^1_{t-k}$ and $\textbf{\textit{M}}^2_{t-k}$ with two different scales are resized to the same size via up-sampling. Then, they are summed element-wise to produce the final motion feature $\hat{\textbf{\textit{M}}}_{t-k}$. Note that  $\hat{\textbf{\textit{M}}}_{t-k}$ has the same size as $\textbf{\textit{F}}_{t-k}$.

 Let  $\Delta \textbf{\textit{P}}_{t-k}$ and  $\textbf{\textit{A}}_{t-k}$ be the offsets and weights of the deformable mask applied to the features $\textbf{\textit{F}}_{t-k}$. Note that $\Delta \textbf{\textit{P}}_{t-k}$ and  $\textbf{\textit{A}}_{t-k}$ are simply obtained by applying a $1 \times 1$ convolution to the motion feature $\hat{\textbf{\textit{M}}}_{t-k}$. Then, the deformable mask with the offsets $\Delta \textbf{\textit{P}}_{t-k}$ and weights $\textbf{\textit{A}}_{t-k}$ is applied to $\textbf{\textit{F}}_{t-k}$ for feature alignment. Finally, this operation produces the aligned BEV feature maps $\{\textbf{\textit{F}}_{k}^{align}\}_{k=t-K+1}^{t-1}$.

\subsection{Spatio-Temporal Feature Aggregation}
STFA produces an enhanced feature map $\hat{\textbf{\textit{F}}}_t$ by aggregating the BEV feature maps $ \textbf{\textit{F}}_{t-K+1}^{align},...,\textbf{\textit{F}}_{t-1}^{align}, \textbf{\textit{F}}_{t}$. We re-design a deformable attention \cite{def-detr} to aggregate multi-frame features with spatio-temporal attention. 

STFA successively decodes the query $\textbf{\textit{Q}}^{(l)}$ over $L$ attention layers, where $l$ denotes the decoding layer index. In the first decoding layer, $\textbf{\textit{Q}}^{(0)}$ is initialized by $\textbf{\textit{F}}_{t}$.
STFA performs simultaneous deformable attention on the $K$ input features $\{\textbf{\textit{X}}_{k}^{(l)}\}_{k=t-K+1}^{t}$ denoted as
\begin{align}
    \textbf{\textit{X}}_{t-k}^{(l)} = \left\{ \begin{array}{ccc} \textbf{\textit{F}}_{t-k}^{align} & \mbox{for} \quad k \neq 0 \\ 
   \textbf{\textit{Q}}^{(l)} &  \mbox{for} \quad k = 0 \end{array} \right. .
\end{align}
 using $K$ {\it derivative queries} $\{\textbf{\textit{H}}_{k}^{(l)}\}_{k=t-K+1}^{t}$.   
The derivative query $\textbf{\textit{H}}_{t-k}^{(l)}$ is derived from the main query $\textbf{\textit{Q}}^{(l)}$ as
\begin{align}
    \textbf{\textit{H}}_{t-k}^{(l)} = \left\{ \begin{array}{ccc} \mathrm{Conv}_{3 \times 3}([\textbf{\textit{F}}_{t-k}^{align}, \textbf{\textit{Q}}^{(l)}]) & \mbox{for} \quad k \neq 0 \\ 
   \textbf{\textit{Q}}^{(l)} &  \mbox{for} \quad k = 0 \end{array} \right. .
\end{align} 
The derivative query $\textbf{\textit{H}}_{t-k}^{(l)}$ is used to determine the offsets and weights of the deformable masks applied to $\textbf{\textit{X}}_{t-k}^{(l)}$. 
When $k \neq 0$, the derivative query $\textbf{\textit{H}}_{t-k}^{(l)}$ depends on both $\textbf{\textit{F}}_{t-k}^{align}$ and $\textbf{\textit{Q}}^{(l)}$, since the deformable mask for $\textbf{\textit{F}}_{t-k}^{align}$ should be determined based on the information present in both the  $(t-k)$th and the $t$th frames. 
In contrast, when $k=0$, the derivative query $\textbf{\textit{H}}_{t}^{(l)}$ is determined solely by $\textbf{\textit{Q}}^{(l)}$.

Figure \ref{long_term_model} (b) depicts the deformable attention with $M$ multi-heads using the derivative queries $\{\textbf{\textit{H}}_{k}^{(l)}\}_{k=t-K+1}^{t}$. 
Let $q$ index $HW$ elements of the input feature map, where $W$ and $H$ denote the width and height of input features. Consider a 2D reference point $p_q$ at the location of $q$th element of the input feature map.
Given the mask offset $\Delta_{t-k,qmj}^{(l)}$ and attention weight $A_{t-k, qmj}^{(l)}$, the feature aggregation is performed as
\begin{align}
    y_q =& \sum_{m=1}^{M}W_m(\sum_{k=0}^{K-1}\sum_{j=1}^{J}A_{t-k, qmj}^{(l)}W_m^{'}  \nonumber \\
    &\cdot \textbf{\textit{X}}_{t-k}^{(l)}(p_q+\Delta_{t-k,qmj}^{(l)})),
\end{align}
where $m$ and $j$ are the multi-head index and sampling point index, respectively. $W_m\in\mathbb{R}^{C \times (C/M)}$ and $W_m^{'}\in\mathbb{R}^{(C/M) \times (C)}$ denote the learnable projection matrices, where $C$ is the channel size of the input feature. We let $\textbf{\textit{X}}_{t-k}^{(l)}(p)$ be the element of $\textbf{\textit{X}}_{t-k}^{(l)}$ at the position $p$. 
The mask offsets $\Delta_{t-k,qm}^{(l)}$ and the attention weights $A_{t-k, qm}^{(l)}$ are obtained from
\begin{align}
    \Delta_{t-k,qm}^{(l)} &= W_{\Delta,m}(\textbf{\textit{H}}_{t-k}^{(l)}(p_q)) \\
    A_{t-k, qm}^{(l)} &= \mathrm{Softmax}(W_{A,m}(\textbf{\textit{H}}_{t-k}^{(l)}(p_q))),
\end{align}
where $\Delta_{t-k,qm}^{(l)}=\left[\Delta_{t-k,qm1}^{(l)},\cdots,\Delta_{t-k,qmJ}^{(l)}\right]^{T}$, and $A_{t-k, qm}^{(l)}=\left[A_{t-k,qm1}^{(l)},\cdots,A_{t-k,qmJ}^{(l)}\right]^{T}$, $\mathrm{Softmax(\cdot)}$ is the softmax function, and $W_{\Delta,m} (\in\mathbb{R}^{2J \times C})$ and $W_{A,m} (\in\mathbb{R}^{J \times C})$ are the learnable projection matrices.

Finally, the $q$th element of main query is updated as
\begin{gather}
    \textbf{\textit{Q}}^{(l+1)}(p_q) = \mathrm{FFN}(\mathrm{LN}(\mathrm{Dropout}(y_q) + \textbf{\textit{Q}}^{(l)}(p_q))),
\end{gather}
where $\mathrm{FFN}(\cdot)$ denotes feed-forward network \cite{attention}, $\mathrm{LN(\cdot)}$ denotes the layer normalization \cite{layernorm}, and $\mathrm{Dropout(\cdot)}$ denotes the drop-out operation \cite{dropout}.
After $L$ layers of query decoding, the aggregated BEV feature maps can finally be obtained as $\hat{\textbf{\textit{F}}}_t = \textbf{\textit{Q}}^{(L)}$.

\newcolumntype{C}{>{\centering\arraybackslash}p{2.3em}}

\renewcommand{\arraystretch}{1.0}

\begin{table*}[t]
\begin{center}

\begin{adjustbox}{width=0.99\linewidth}

\begin{tabular}{c || C  C |  C  C  C  C  C  C  C  C  C  C }
\Xhline{4\arrayrulewidth}

Method & NDS & mAP & Car & Truck
& Bus & Trailer & C.V & Ped. & Motor. & Bicycle & T.C & Barrier \\ \hline\hline

PointPillars \cite{pointpillars} & 45.3 & 30.5 & 68.4 & 23.0 & 28.2 & 23.4 & 4.1 & 59.7 & 27.4 & 1.1 & 30.8 & 38.9\\
WYSIWYG \cite{wysiwyg} & 41.9 & 35.0 & 79.1 & 30.4 & 46.6 & 40.1 & 7.1 & 65.0 & 18.2 & 0.1 & 28.8 & 34.7\\
3DSSD \cite{3dssd} & 56.4 & 42.6 & 81.2 & 47.2 & 61.4 & 30.5 & 12.6 & 7.2 & 36.0 & 8.6 & 31.1 & 47.9\\
SA-Det3D \cite{sa-det3d} & 59.2 & 47.0 & 81.2 & 43.8 & 57.2 & 47.8 & 11.3 & 73.3 & 32.1 & 7.9 & 60.6 & 55.3\\
SSN V2 \cite{ssnV2} & 61.6 & 50.6 & 82.4 & 41.8 & 46.1 & 48.0 & 17.5 & 75.6 & 48.9 & 24.6 & 60.1 & 61.2\\
CBGS \cite{cbgs} & 63.3 & 52.8 & 81.1 & 48.5 & 54.9 & 42.9 & 10.5 & 80.1 & 51.5 & 22.3 & 70.9 & 65.7\\
CVCNet \cite{cvcnet} & 64.2 & 55.8 & 82.7 & 46.1 & 45.8 & 46.7 & 20.7 & 81.0 & 61.3 & 34.3 & 69.7 & 69.9\\
HotSpotNet \cite{hotspotnet} & 66.6 & 59.3 & 83.1 & 50.9 & 56.4 & 53.3 & 23.0 & 81.3 & 63.5 & 36.6 & 73.0 & 71.6\\
CyliNet \cite{cylinet} & 66.1 & 58.5 & 85.0 & 50.2 & 56.9 & 52.6 & 19.1 & 84.3 & 58.6 & 29.8 & 79.1 & 69.0\\
CenterPoint \cite{centerpoint} & 67.3 & 60.3 & 85.2 & 53.5 & 63.6 & 56.0 & 20.0 & 84.6 & 59.5 & 30.7 & 78.4 & 71.1\\
AFDetV2 \cite{afdetv2} & 68.5 & 62.4 & 86.3 & 54.2 & 62.5 & 58.9 & 26.7 & 85.8 & 63.8 & 34.3 & 80.1 & 71.0 \\
S2M2-SSD \cite{s2m2ssd} & 69.3 & 62.9 & 86.3 & 56.0 & 65.4 & 59.8 & 26.2 & 84.5 & 61.6 & 36.4 & 77.7 & 75.1\\
TransFusion-L \cite{transfusion} & 70.2 & 65.5 & 86.2 & 56.7 & 66.3 & 58.8 & 28.2 & 86.1 & 68.3 & 44.2 & 82.0 & \textbf{78.2}\\
VISTA \cite{vista} & 70.4 & 63.7 & 84.7 & 54.2 & 64.0 & 55.0 & 29.1 & 83.6 & 71.0 & 45.2 & 78.6 & 71.8\\

\hline

3DVID (with PointPillars) \cite{3dvid}  & 53.1 & 45.4 & 79.7 & 33.6 & 47.1 & 43.1 & 18.1 & 76.5 & 40.7 & 7.9 & 58.8 & 48.8\\
3DVID (with CenterPoint) \cite{3DVID_TPAMI}  & 71.4 & 65.4 & 87.5 & 56.9 & 63.5 & 60.2 & \textbf{32.1} &  82.1 &  74.6 & 45.9 & 78.8 & 69.3\\
MGTANet-P  & 61.4 & 50.9 & 81.3 & 45.8 & 55.0 & 48.9 & 18.2 & 74.4 & 52.6 & 17.8 & 61.7 & 53.0\\
MGTANet-C  & 71.2 & 65.4 & 87.7 & 56.9 & 64.6 & 59.0 & 28.5 & 86.4 & 72.7 & 47.9 & 83.8 & 65.9\\
MGTANet-CT  &  \textbf{72.7} &  \textbf{67.5} &  \textbf{88.5} &  \textbf{59.8} &  \textbf{67.2} &  \textbf{61.5} & 30.6 & \textbf{87.3} & \textbf{75.8} &  \textbf{52.5} &  \textbf{85.5} & 66.3\\

\Xhline{4\arrayrulewidth}

\end{tabular}
\end{adjustbox}
\end{center}

\caption{\textbf{Performance on nuScenes \textit{test} benchmark.} The model is trained on nuScenes \textit{train} set and evaluated on nuScenes \textit{test} set. C.V and T.C, respectively, indicates the construction vehicle and the traffic cone. Ped. and Motor. are short for the motorcycle and the pedestrian, respectively. The best performance are in boldface.}

\label{table:sota}
\end{table*}
\renewcommand{\arraystretch}{1}

\section{Experiments}
\subsection{nuScenes dataset}
The nuScenes dataset \cite{nuscenes} is a large-scale autonomous driving dataset that contains 700, 150, and 150 scenes for training, validation, and testing. It comprises LiDAR point cloud data acquired at 20Hz using 32-channel LiDAR. Keyframe samples are provided at 2Hz with 360-degree object annotations. The dataset also provides intermediate sensor frames. Average precision (AP) metric defines a match by thresholding the 2D center distance on the ground plane, and NDS metric is the one suggested by nuScenes 3D object detection benchmark \cite{nuscenes}.  These metrics were used to evaluate the performance of our proposed 3D detection method on 10 object categories (barrier, bicycle, bus, car, motorcycle, pedestrian, trailer, truck, construction vehicle, and traffic cone).

\subsection{Implementation details}
{\bf Data Processing.} Following the format of nuScenes dataset \cite{nuscenes}, a single frame consists of $N=10$ multiple LiDAR scans, i.e., 10 point sets. Each LiDAR point is represented as $(x,y,z,r,\Delta t)$, where $(x,y,z)$ denotes the coordinate of a 3D point, $r$ denotes the reflectance, and $\Delta t$ denotes the time delta in seconds from the keyframe. In our experiment, the number of frames  $K$ processed by the MGTANet was set to 3, corresponding to a duration of 1.5 seconds. 

\noindent
\textbf{Network Architecture.}
The proposed method was implemented on two widely used 3D object detectors, PointPillars \cite{pointpillars} and CenterPoint \cite{centerpoint}. For PointPillars, the range of point clouds was within $[-51.2,51.2]\times[-51.2,51.2]\times[-5.0,3.0]m$ on the $(x,y,z)$ axes and the LiDAR voxel structure comprised $512 \times 512 \times 1$ voxel grids with a grid size of $0.2 \times 0.2 \times 8.0m$. We chose the CenterPoint \cite{centerpoint} using the backbone used in SECOND \cite{second}. The voxelization range was set to $[-54.0,54.0]\times[-54.0,54.0]\times[-5.0,3.0]m$ along the $(x,y,z)$ axes. The size of a voxel was set to $(0.075 \times 0.075 \times 0.2m)$, and the dimension of the voxel structure was set to $1440 \times 1440 \times 40$.
In our evaluation, we considered three versions of MGTANet, including 1) {\it MGTANet-P}: MGTANet with a PointPillars baseline, 2) {\it MGTANet-C}: MGTANet with a CenterPoint baseline, and 3) {\it MGTANet-CT}: MGTANet with a CenterPoint baseline and with test time augmentation \cite{tta} enabled. Following the configuration of the 3DVID method \cite{3DVID_TPAMI}, MGTANet-C and MGTANet-CT were implemented to allow a latency by one frame.

\noindent
{\bf Training.} Our model was trained in two stages. In the first stage, we trained the SM-VFE, backbone, and detection head jointly in the same manner as single-frame 3D detectors. A one-cycle learning rate policy was used for 20 epochs with a maximum learning rate of 0.001. After initializing the model with the weights obtained from the first training stage, our entire network was fine-tuned for 40 epochs with the same learning rate scheduling policy but with a maximum learning rate of 0.0002. The Adam optimizer was used to optimize the network in both stages. We utilized the loss function used in \cite{pointpillars} and \cite{centerpoint}. We applied data augmentation methods during pre-training and fine-tuning, including random flipping, rotation, scaling, and ground-truth box sampling \cite{second}. The batch size was set to 16 for MGTANet-P and 8 for MGTANet-C and MGTANet-CT.

\newcolumntype{D}{>{\centering\arraybackslash}p{9.5em}}
\newcolumntype{M}{>{\centering\arraybackslash}p{8.0em}}
\newcolumntype{A}{>{\centering\arraybackslash}p{4.5em}}
\renewcommand{\arraystretch}{1.2}
\begin{table*}[t]
\begin{center}
\begin{adjustbox}{width=0.92\textwidth}
\begin{tabular}{c||D|D|D||A|A}
\Xhline{4\arrayrulewidth}
 \multirow{3}{*}{{\it Method}} & \multicolumn{3}{c||}{{\it Proposed 3D-PCS Strategy}} &\multicolumn{2}{c}{Performance}\\ \cline{2-6}
 &\multirow{2}{*}{\begin{tabular}[c]{@{}c@{}}Short-term \\ Motion-aware VFE \end{tabular}}&
 \multirow{2}{*}{\begin{tabular}[c]{@{}c@{}}Spatio-Temporal\\Feature Aggregation\end{tabular}}& \multirow{2}{*}{\begin{tabular}[c]{@{}c@{}}Motion-Guided\\Deformable Alignment\end{tabular}} & 
 \multirow{2}{*}{mAP (\%)} & \multirow{2}{*}{NDS (\%)}\\ 
 &&&&&\\
 \hline\hline
 Baseline & & & & 54.99 & 63.33\\
\hline 

\multirow{3}{*}{\begin{tabular}[c]{@{}c@{}}
Our\\MGTANet\end{tabular}} & \checkmark &  & 
& 56.52$_{\uparrow 1.53}$ & 64.22$_{\uparrow 0.89}$  \\
& \checkmark & \checkmark  & & 58.24$_{\uparrow 3.25}$  & 65.22$_{\uparrow 1.89}$  \\
& \checkmark & \checkmark & \checkmark & {59.61$_{\uparrow 4.62}$} & {65.94$_{\uparrow 2.61}$}  \\

\Xhline{4\arrayrulewidth}
\end{tabular}
\end{adjustbox}
\end{center}
\caption{{\bf Ablation study on the nuScenes {\it valid} set. } The effects of three main modules are highlighted in this study.}
\label{table:ablation}
\end{table*}
\renewcommand{\arraystretch}{1}
\newcolumntype{E}{>{\centering\arraybackslash}p{10.6em}}
\renewcommand{\arraystretch}{1.2}

\begin{table}[t]
\begin{center}
\begin{adjustbox}{width=0.42\textwidth}
\begin{tabular}{E ||c|c}

\Xhline{4\arrayrulewidth}
\multirow{2}{*}{{\it Method}} & \multicolumn{2}{c}{Performance}\\ \cline{2-3}
& mAP (\%) & NDS (\%) \\ \hline \hline

Baseline & 54.99 & 63.33 \\ \hline

+ Motion embedding & 56.24$_{\uparrow 1.25}$ & 64.00$_{\uparrow 0.67}$ \\ 

+ Channel-wise attention & 56.52$_{\uparrow 1.53}$ & 64.22$_{\uparrow 0.89}$ \\ 

\Xhline{4\arrayrulewidth}

\end{tabular}
\end{adjustbox}
\caption{{\bf Ablation study to evaluate the sub-modules of SM-VFE.}}
\label{table:smvfe}
\end{center}
\end{table}

\renewcommand{\arraystretch}{1}

\subsection{Performance on nuScenes {\it test} set.} 
Table \ref{table:sota} provides the performance of several LiDAR-based 3D object detectors evaluated on nuScenes 3D object detection tasks. The results for other LiDAR-based methods are brought from the nuScenes leaderboard\footnote{https://www.nuscenes.org/object-detection?externalData=all\&mapData=all\&modalities=Lidar} except for 3DVID\footnote{The performance of 3DVID on the nuScenes learderboard was obtained with PointPainting sensor fusion \cite{pointpainting}  enabled. For a fair comparison, we added the performance in the original papers \cite{3dvid, 3DVID_TPAMI} to the table.}.   Note that  MGTANet-CT outperforms other latest 3D object detectors in the leaderboard. To the best of our knowledge, 3DVID \cite{3dvid, 3DVID_TPAMI} had the record of state-of-the-art (SOTA) performance among the existing methods based on point cloud sequences. MGTANet-CT surpassed 3DVID, setting a new SOTA performance.  Note that MGTANet-P achieved an 8.3\% better performance in NDS compared to the 3DVID with PointPillars baseline \cite{3dvid}. The performance of MGTANet-C is comparable to that of 3DVID with  a CenterPoint baseline \cite{3DVID_TPAMI}; however, MGTANet-CT demonstrates better performance.

By encoding point cloud sequences, MGTANet exhibited a remarkable improvement in performance compared to the baseline methods. MGTANet-P achieved 16.1\% gain in NDS and 20.4\% gain in mAP over PointPillars baseline. 
MGTANet-C achieved up to 3.9\% gain in NDS and 5.1\% gain in mAP over CenterPoint baseline.
This demonstrates that the spatio-temporal context information contained in a point cloud sequence can significantly contribute to improving the performance of 3D object detection method.

\subsection{Ablation studies on nuScenes {\it valid} set}
We conducted several ablation studies on the nuScenes validation set. To reduce the time required for these experiments, we used only 1/7 of the training set to perform training and the entire validation set for evaluation. We used CenterPoint \cite{centerpoint} as a baseline detector for all ablation studies.

\noindent
{\bf Three main modules.} Table \ref{table:ablation} shows the contributions of SM-VFE, MGDA, and STFA to the overall performance. Using spatio-temporal information in the voxel encoding stage, SM-VFE improves the performance of the baseline by 1.53\% in mAP. When both SM-VFE and STFA are enabled, i.e., the multiple BEV features encoded using SM-VFE are aggregated by STFA without feature alignment, the mAP performance is improved by additional 1.72\%.  Finally, when we add MGDA module to boost the effect of feature aggregation, 1.37\% further mAP improvement is achieved. 
Combination of all three modules offers  a total performance gain of 4.62\% in mAP and  2.61\% in NDS over the CenterPoint baseline.

\noindent
{\bf Sub-modules of SM-VFE.} Table \ref{table:smvfe} provides the performance achieved by each component of SM-VFE. We use the vanilla voxel encoding method of CenterPoint as a baseline. When we add the motion embedding to the voxel features, the mAP is improved by 1.25\% over the baseline. The channel-wise attention strategy yields additional gains of 0.28\% in mAP and 0.22\% in NDS by weighting the motion-sensitive channels of the latent features.

\noindent
{\bf Derivative query for STFA.} 
Table \ref{table:stfa} demonstrates the effectiveness of the derivative queries used in STFA.
Baseline* indicates the method that adds SM-VFE voxel encoding method to CenterPoint.
{\it Single query} indicates the method that determines the deformable masks only using the main query features.
Note that the derivative queries offer a performance gain of 0.74\% in mAP and 0.38\% in NDS over the single query.  
This shows that the derivative queries effectively leverage the capabilities of our spatio-temporal cross-attention mechanism.

\newcolumntype{F}{>{\centering\arraybackslash}p{4.7em}}
\renewcommand{\arraystretch}{1.2}

\begin{table}[t]
\begin{center}
\begin{adjustbox}{width=0.42\textwidth}
\begin{tabular}{F | F || c | c}

\Xhline{4\arrayrulewidth}
\multirow{2}{*}{{\it Method}} & \multirow{2}{*}{{Query type}} & \multicolumn{2}{c}{Performance}\\ \cline{3-4}
& & mAP (\%) & NDS (\%) \\ \hline \hline
Baseline* &        -          & 56.52        & 64.22        \\ \hline
\multirow{2}{*}{STFA} & Single & 57.50$_{\uparrow 0.98}$ & 64.84$_{\uparrow 0.62}$ \\
                      & Derivative & 58.24$_{\uparrow 1.72}$ & 65.22$_{\uparrow 1.00}$ \\
\Xhline{4\arrayrulewidth}      




\end{tabular}
\end{adjustbox}
\caption{\textbf{Comparison of derivative queries versus a single query in STFA.}}
\label{table:stfa}
\end{center}
\end{table}

\renewcommand{\arraystretch}{1}

\section{Conclusions}
In this paper, we proposed a new 3D object detection method MGTANet designed to model temporal structures in point cloud sequences. The proposed MGTANet can effectively find the spatio-temporal representation of point cloud sequences using motion context. First, we proposed the enhanced voxel encoding method, SM-VFE to model the temporal distribution of points acquired from consecutive LiDAR scans. We devised the latent motion embedding method that can enhance the quality of the voxel features. Second, we proposed the architecture for aggregating the BEV feature maps produced by the SM-VFE  over multiple frames. First, MGDA  aligned the adjacent BEV feature maps through the deformable convolution. The offsets and weights of the deformable masks were determined based on multi-scale motion features. STFA then aggregated the multiple BEV feature maps aligned by MGDA using  spatio-temporal deformable attention. We introduced the derivative queries, which can enable simultaneous co-attention to the adjacent BEV features. Our evaluation conducted on nuScenes dataset confirmed that the proposed MGTANet exhibited significant improvements in performance compared to LiDAR-only baselines; moreover it outperformed the latest top-ranked 3D-PCS methods on the nuScenes 3D object detection leaderboard.
 
\newpage

\section{Acknowledgements}
This work was partly supported by 1) Institute of Information \& communications Technology Planning \& Evaluation (IITP) grant funded by the Korea government (MSIT) (No.2020-0-01373, Artificial Intelligence Graduate School Program(Hanyang University)), 2) National Research Foundation of Korea (NRF) grant funded by the Korea government (MSIT) (No.2020R1A2C2012146), and 3) Institute for Information \& communications Technology Promotion (IITP) grant funded by the Korea government (MSIP) (No.2021-0-01314,Development of driving environment data stitching technology to provide data on shaded areas for autonomous vehicles)

\bibliography{aaai23}

\newpage

\onecolumn
\renewcommand{\thesection}{\Alph{section}}
\setcounter{figure}{0}
\setcounter{table}{0}
\section{Appendix}
\begin{figure*}[h]
    \centerline{\includegraphics[width=0.78\textwidth]{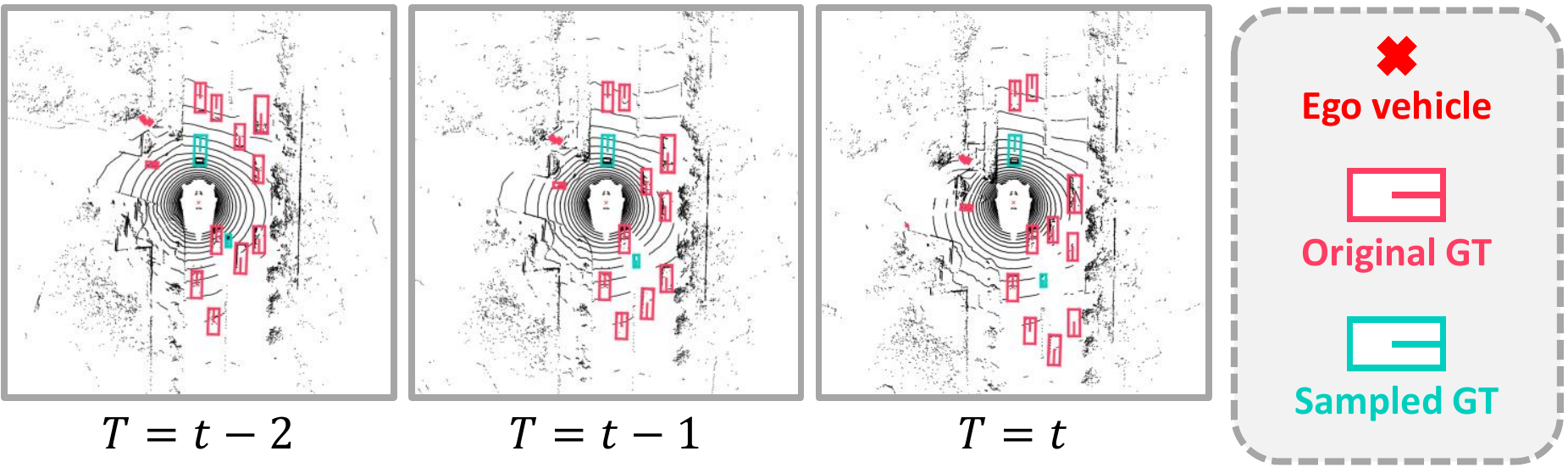}}
	\caption {\textbf{Visualization of GT sampling.} We implement GT sampling by keeping relative motion of objects within sequence.}
	\label{seqgt}
\end{figure*}
\subsection{Implementation Details of Data Augmentations}
In training stage, we apply the data augmentations, including random flipping, global scaling, global rotation, and GT sampling\cite{second}. We randomly flipped the LiDAR points, and rotated the point cloud within a range of $[-\frac{\pi}{4},\frac{\pi}{4}]$ along the $z$ axis. We also scaled the coordinates of the points with a factor within $[0.95.1.05]$. 
As used in \cite{second}, we randomly sampled several ground truth 3D boxes from other scenes and pasted corresponding 3D objects as a point cloud into the frame of interest. To leverage the capabilities of temporal information, this augmentation was applied to the neighboring frame while maintaining the relative motion pattern within sequence as depicted in Figure \ref{seqgt}.
In the last 5 epochs in training stage, we apply these augmentation techniques except GT sampling to point cloud sequences by consistently using random parameters.

\subsection{Additional Experiments}
\noindent
\textbf{Effectiveness of MGDA. }
Table \ref{table:mgda} shows the effectiveness of MGDA. To simplify these experiments, BEV features are aggregated by the simple method, where the BEV features from multiple frames are concatenated, and fed into a $3\times3$ convolution layer. When this simple strategy is applied to the CenterPoint without MGDA, the performance is reduced by 2.36\% in mAP and by 1.80\% in NDS. In contrast, we observe that MGDA improves the performance by 3.38\% in mAP and by 2.26\% in NDS over baseline. This demonstrates that feature alignment is considered crucial for performing feature aggregation.

\newcolumntype{S}{>{\centering\arraybackslash}p{9.5em}}
\renewcommand{\arraystretch}{1.2}

\begin{table}[h]
\centering
\begin{adjustbox}{width=0.40\textwidth}

\begin{tabular}{S ||c|c}

\Xhline{4\arrayrulewidth}
\multirow{2}{*}{{\it Method}} & \multicolumn{2}{c}{Performance}\\ \cline{2-3}
& mAP (\%) & NDS (\%) \\ \hline \hline
Baseline                  & 54.99        & 63.33        \\ \hline
w/o MGDA & 52.63 ($\downarrow 2.36$) & 61.53 ($\downarrow 1.80$) \\ 
with MGDA & 58.37 ($\uparrow 3.38$) & 65.59 ($\uparrow 2.26$) \\ \hline

\Xhline{4\arrayrulewidth}

\end{tabular}
\end{adjustbox}
\caption{{\bf Ablation study for effectiveness of MGDA.}}
\label{table:mgda}
\end{table}
\renewcommand{\arraystretch}{1}

\noindent
\textbf{Performance comparison on nuScenes} \textit{valid} \textbf{set}
Table \ref{table:valid} compares our MGTANet with several LiDAR-based 3D object detectors on the nuScenes\cite{nuscenes}. We used the whole training set for training and the whole validation set for evaluating. Our framework works in both online and offline modes. The MGTANet-C with online mode employs two previous frames to produce detection results in the frame of interest, and achieves 3.3\% performance gain in mAP and 1.9\% performance gain in NDS over our baseline. When applying offline mode using a previous frame and a current frame, our model achieves the best results with 64.8\% mAP and 70.6\% NDS.

\newcolumntype{J}{>{\centering\arraybackslash}p{2.3em}}

\renewcommand{\arraystretch}{1.0}

\begin{table*}[h]
\begin{center}

\begin{adjustbox}{width=0.95\textwidth}

\begin{tabular}{c || J  J |  J  J  J  J  J  J  J  J  J  J }
\Xhline{4\arrayrulewidth}


Method & NDS & mAP & Car & Truck
& Bus & Trailer & C.V & Ped. & Motor. & Bicycle & T.C & Barrier \\ \hline\hline


CBGS \cite{cbgs} & 56.3 & 56.3 & 82.9 & 52.9 & 64.7 & 37.5 & 18.3 & 80.3 & 60.1 & 39.4 & 64.8 & 64.3\\
CVCNet \cite{cvcnet} & 65.5 & 54.6 & 83.2 & 50.0 & 62.0 & 34.5 & 20.2 & 81.2 & 54.4 & 33.9 & 61.1 & 65.5\\
HotSpotNet \cite{hotspotnet} & 66.0 & 59.5 & 84.0 & 56.2 & 67.4 & 38.0 & 20.7 & 82.6 & 66.2 & 49.7 & 65.8 & 64.3\\
CenterPoint* \cite{centerpoint} & 66.8 & 59.6 & 85.5 & 58.6 & 71.5 & 37.3 & 17.1 & 85.1 & 58.9 & 43.4 & 69.7 & \textbf{68.5}\\


\hline
MGTANet-C* (Online-mode)  & 68.7 & 62.9 & 87.0 & 59.6 & 72.3 & 40.1 & 21.5 & 86.3 & 69.3 & 51.4 & 73.4 & 67.8\\
MGTANet-C* (Offline-mode)  & \textbf{70.6} & \textbf{64.8} & \textbf{87.9} & \textbf{61.3} & \textbf{73.2} & \textbf{40.2} & \textbf{23.0} & \textbf{86.6} & \textbf{74.1} & \textbf{59.9} & \textbf{75.7} & 66.1\\

\Xhline{4\arrayrulewidth}

\end{tabular}
\end{adjustbox}
\end{center}
\caption{\textbf{Performance on nuScenes} \textit{valid} \textbf{set.} The model is trained on nuScenes whole \textit{train} set and evaluated on nuScenes \textit{val} set. * denotes the model implemented in the released code of CenterPoint\cite{centerpoint}.}

\label{table:valid}
\end{table*}
\renewcommand{\arraystretch}{1}

\begin{figure*}[h]
    \centering
    \begin{subfigure}[]
    {
        \includegraphics[width=0.74\textwidth]{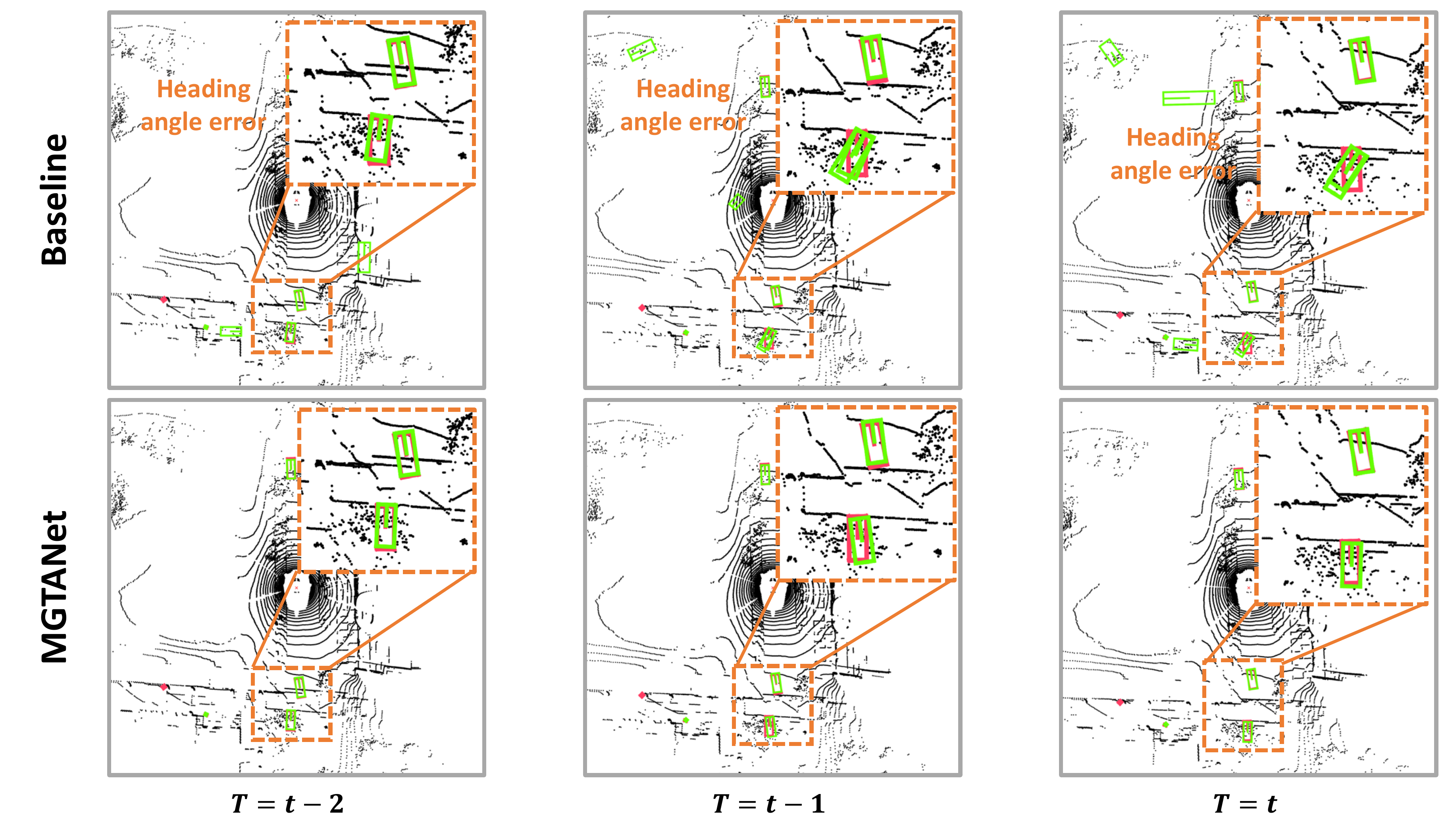}
    }
    \end{subfigure}
    \begin{subfigure}[]
    {
        \includegraphics[width=0.74\textwidth]{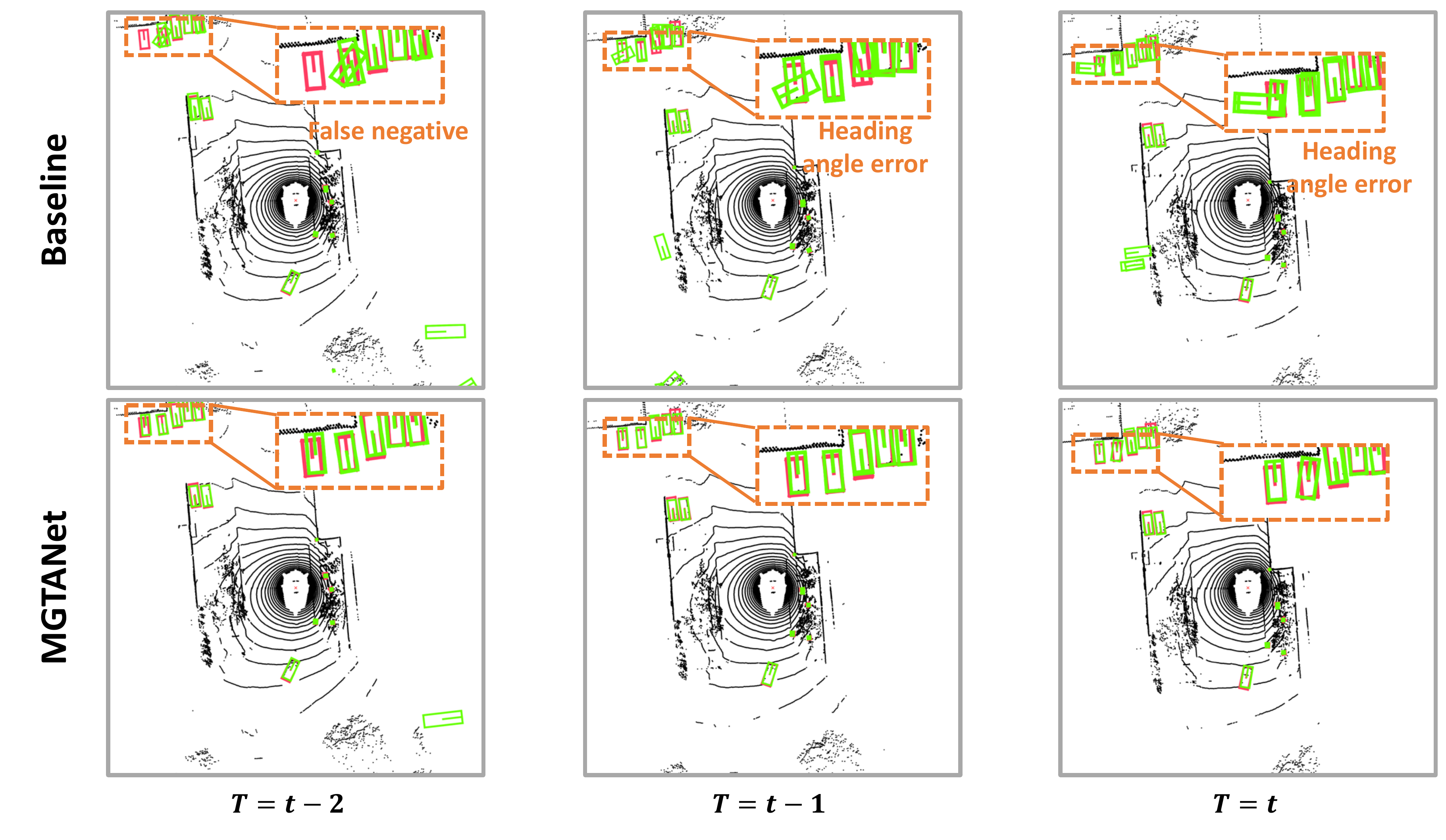}
    }
    \end{subfigure}
    \hspace{0cm}
    \caption {\textbf{Comparing of CenterPoint and our MGTANet.} Predicted and ground truth 3D bounding boxes are assigned with green and red, respectively. It can be found that our method shows much better results when objects are too far.
    }
    \label{visualization_far}
\end{figure*}

\begin{figure*}[t]
    \centering
    \begin{subfigure}[]
    {
        \includegraphics[width=0.74\textwidth]{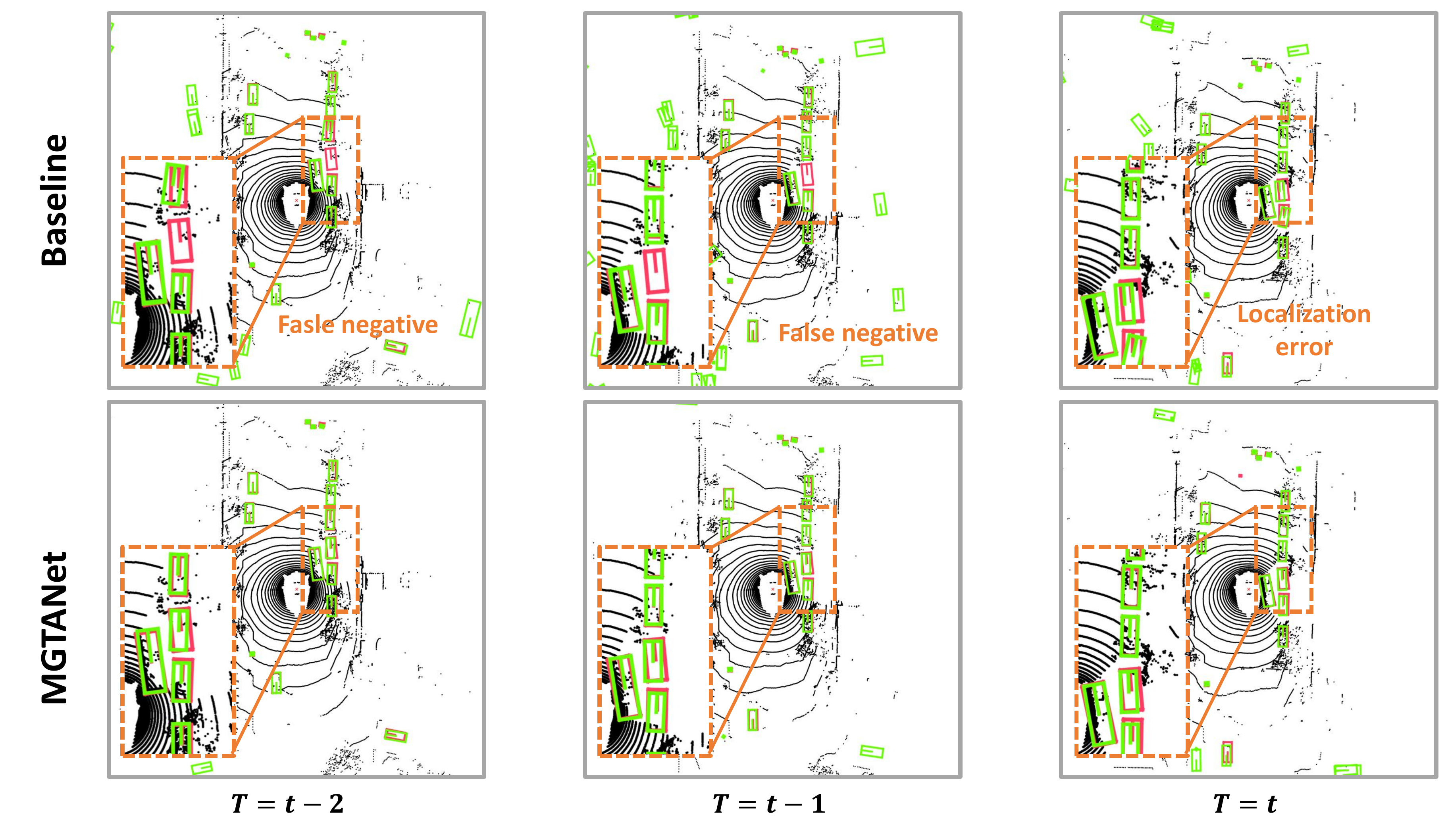}
    }
    \end{subfigure}
    \hspace{0cm}
    \begin{subfigure}[]
    {
        \includegraphics[width=0.74\textwidth]{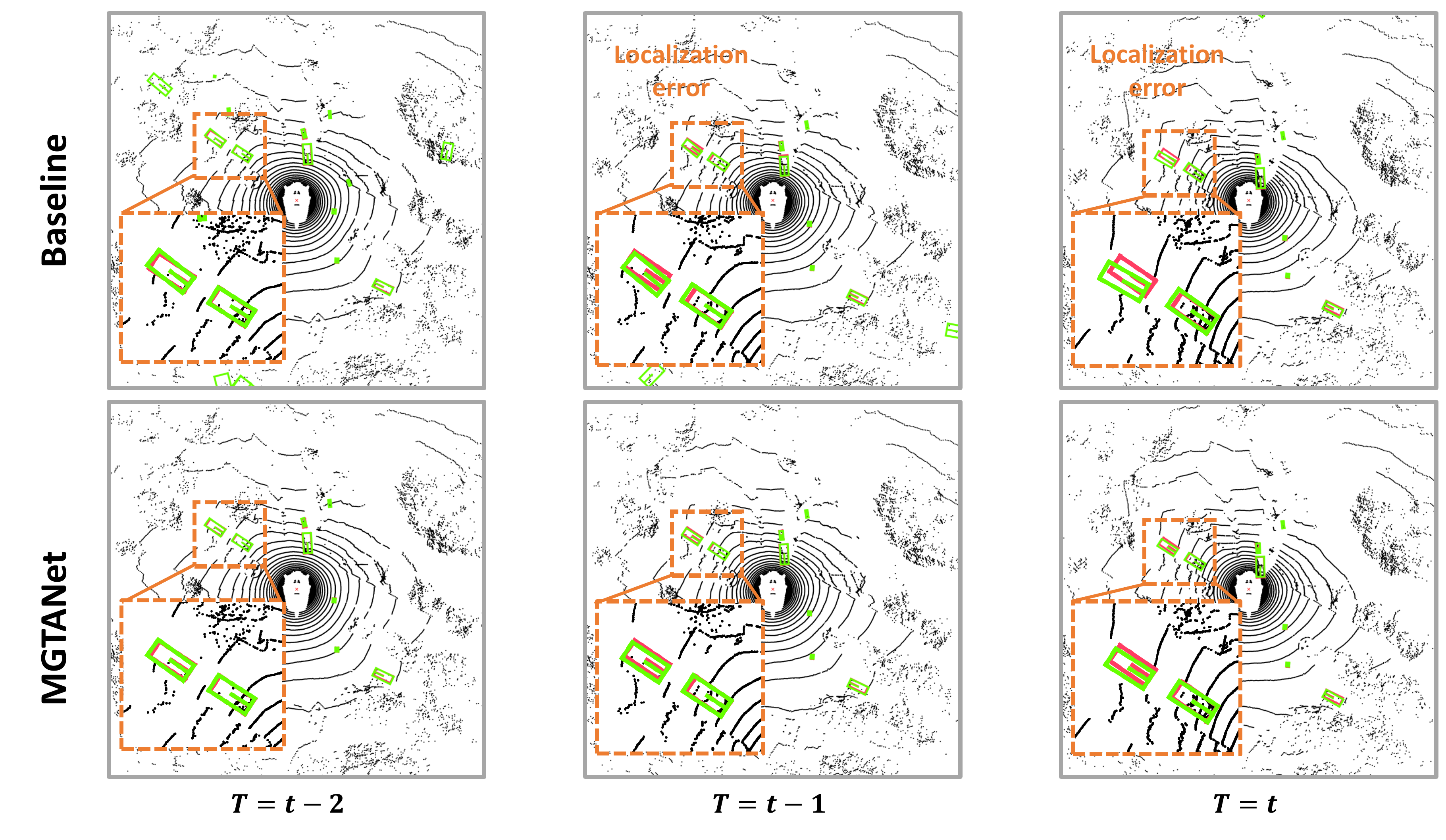}
    }
    \end{subfigure}
    \hspace{0cm}
    \caption {\textbf{Comparing of CenterPoint and our MGTANet.} Predicted and ground truth 3D bounding boxes are assigned with green and red, respectively. It can be found that our method shows much better results when objects are occluded by the others.
    }
    \label{visualization_occ}
\end{figure*}

\noindent
\textbf{Qualitative Results. }
Figure \ref{visualization_far} and \ref{visualization_occ} shows the qualitative results of our MGTANet and CenterPoint. Our model can produce more accurate 3D detection results with fewer false positives than CenterPoint. In Figure \ref{visualization_far}, CenterPoint fails to detect far distant objects whose point clouds are sparse. In contrast, MGTANet accurately recognize these objects thanks to the temporal information from adjacent frames. 
Similarly, when partial occlusion occurs, as shown in Figure \ref{visualization_occ}, MGTANet detects objects more accurately than CenterPoint.

\end{document}